\pdfoutput=1
\documentclass[11pt,a4paper]{article}
\usepackage[hyperref]{acl2020}
\usepackage{nert}
\usepackage{times}
\usepackage{latexsym}

\usepackage{microtype}

\usepackage{booktabs}

\usepackage{xcolor}

\usepackage{devanagari}

\usepackage{textcomp} 
\usepackage{tipa}
\newcommand{\ipa}{\textipa}
\usepackage{url}
\usepackage{graphicx}

\newcommand{\ortho}[1]{$\langle$#1$\rangle$}
\newcommand{\oipa}[1]{{\fontfamily{cmr}\ortho{\ipa{#1}}}}

\aclfinalcopy 
\def\aclpaperid{2133} 

\title{Supervised Grapheme-to-Phoneme Conversion of Orthographic Schwas in Hindi and Punjabi}

\author{Aryaman Arora\textsuperscript{*,\dagger} \quad Luke Gessler\textsuperscript{\dagger} \quad Nathan Schneider\textsuperscript{\dagger} \\
  \textsuperscript{*}School Without Walls High School \\
  \textsuperscript{\dagger}Georgetown University \\
  \texttt{\{\emldisplay{aa2190@georgetown.edu}{aa2190},
  \emldisplay{lg876@georgetown.edu}{lg876},
  \emldisplay{nathan.schneider@georgetown.edu}{nathan.schneider}\}@georgetown.edu} \\}

\date{}

\begin{document}

\renewcommand\textipa[1]{{\fontfamily{cmr}\tipaencoding #1}}

\maketitle

\begin{abstract}
Hindi grapheme-to-phoneme (G2P) conversion is mostly trivial, with one exception: whether a schwa represented in the orthography is pronounced or unpronounced (deleted). Previous work has attempted to predict schwa deletion in a rule-based fashion using prosodic or phonetic analysis. 
We present the first statistical schwa deletion classifier for Hindi, which relies solely on the orthography as the input and outperforms previous approaches. 
We trained our model on a newly-compiled 
pronunciation lexicon
extracted from various online dictionaries.
Our best Hindi model achieves state of the art performance, and also achieves good performance on a closely related language, Punjabi, without modification. 
\end{abstract}

\section{Introduction}

Hindi is written in the Devanagari script, which is an abugida, an orthographic system where the basic unit consists of a consonant and an optional vowel diacritic or a single vowel. Devanagari is fairly regular, but a Hindi word's actual pronunciation can differ from what is literally written in the Devanagari script.\footnote{Throughout this paper, we will adopt the convention of using $\langle$angle brackets$\rangle$ to describe how a word is literally spelled, and [square brackets] to describe how a word is actually pronounced.}
 For instance, in the Hindi word
{\dn p\?pr} $\langle$\ipa{pep@R@}$\rangle$ `paper',
there are three units 
{\dn p\?}~$\langle$\ipa{pe}$\rangle$,
{\dn p}~$\langle$\ipa{p@}$\rangle$, and
{\dn r}~$\langle$\ipa{R@}$\rangle$,
corresponding to the pronounced forms \ipa{[pe]}, \ipa{[p@]}, and \ipa{[r]}. The second unit's inherent schwa is retained in the pronounced form, but the third unit's inherent schwa is deleted.

Predicting whether a schwa will be deleted from a word's orthographic form is generally difficult. Some reliable rules can be stated, e.g. `delete any schwa at the end of the word', but these do not perform well enough for use in an application that requires schwa deletion, like a text-to-speech synthesis system.

This work approaches the problem of predicting schwa deletion in Hindi with machine learning techniques, achieving high accuracy with minimal human intervention. We also successfully apply our Hindi schwa deletion model to a related language, Punjabi. Our scripts for obtaining machine-readable versions of the Hindi and Punjabi pronunciation datasets are published to facilitate future comparisons.\footnote{All of the code, models, and datasets for this research are publicly available at \url{https://github.com/aryamanarora/schwa-deletion}.}

\section{Previous Work}
Previous approaches to schwa deletion in Hindi broadly fall into two classes. 

The first class is characterized by its use of rules given in the formalism of {\it The Sound Pattern of English} \citep{spe}. Looking to analyses of schwa deletion produced by linguists \citep[e.g.,][]{ohala_1983} in this framework, others built schwa deletion systems by implementing their rules. For example, this is a rule used by \citet{narasimhan_schwa-deletion_2004}, describing schwa deletion for words like {\dn j\2glF} $\langle$\ipa{{dZ}@Ng@li:}$\rangle$:

\vspace{0.4em}
\begin{center}
\begin{tabular}{cccccccccccccc}
    C & V & C & C & \textbf{a} & C & V & & C & V & C & C & C & V \\
    \ipa{{dZ}} & \ipa{@} & \ipa{N} & \ipa{g} & \textbf{\ipa{@}} & \ipa{l} & \ipa{i:} & $\rightarrow$ & \ipa{{dZ}} & \ipa{@} & \ipa{N} & \ipa{g} & \ipa{l} & \ipa{i:}
\end{tabular}
\end{center}
\vspace{0.4em}

\noindent Paraphrasing, this rule could be read, ``if a schwa occurs with a vowel and two consonants to its left, and a consonant and a vowel to its right, it should be deleted.'' A typical system of this class would apply many of these rules to reach a word's output form, sometimes along with other information, like the set of allowable consonant clusters in Hindi. These systems were able to achieve fair accuracy (\citeauthor{narasimhan_schwa-deletion_2004}\ achieve 89\%), but were ill-equipped to deal with cases that seemed to rely on detailed facts about Hindi morphology and prosody.

\begin{figure}
    \centering
    \includegraphics[width=0.5\linewidth]{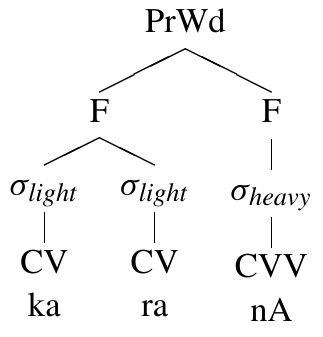}
    \caption{A representative example of the linguistic representations used by \citeauthor{tyson_prosodic_2009} \citeyearpar{tyson_prosodic_2009}. Proceeding from top to bottom, a prosodic word (PrWd) consists of feet, syllables (which have weights), and syllable templates.}
    \label{fig:pstructure}
\end{figure}

Systems of the second class make use of linguistically richer representations of words. Typical of this class is the system of \citet{tyson_prosodic_2009}, which analyzes each word into a hierarchical phonological representation (see figure \ref{fig:pstructure}). These same representations had been used in linguistic analyses: \citet{pandey}, for instance, as noted by \citet{tyson_prosodic_2009}, ``claimed that schwas in Hindi cannot appear between a strong and weak rhyme\footnote{The {\it rhyme} in Hindi (not pictured in \cref{fig:pstructure}), is the part of the syllable that begins with the vowel and includes any consonants that come after the vowel. Its weight is determined by vowel length and whether any consonants appear in it.} within a prosodic foot.'' Systems using prosodic representations perform fairly well, with \citeposs{tyson_prosodic_2009} system achieving performance ranging from 86\% to 94\% but prosody proved not to be a silver bullet; \citet{tyson_prosodic_2009} remark, ``it appears that schwa deletion is a phenomenon governed by not only prosodic information but by the observance of the phonotactics of consonant clusters.''

There are other approaches to subsets of the schwa-deletion problem. One is the diachronic analysis applied by \citet{choudhury} which achieved 99.80\% word-level accuracy on native Sanskrit-derived terms.

Machine learning has not been applied to schwa deletion in Hindi prior to our work. \citet{johny_brahmic_2018} used neural networks to model schwa deletion in Bengali (which is not a binary classification problem as in Hindi) and achieved great advances in accuracy. We employ a similar approach to Hindi, but go further by applying gradient-boosting decision trees to the problem, which are more easily interpreted in a linguistic format.

Similar research has been undertaken in other Indo-Aryan languages that undergo schwa-deletion, albeit to a lesser extent than Hindi. \citet{wasala-06}, for example, proposed a rigorous rule-based G2P system for Sinhala.

\section{Methodology}

We frame schwa deletion as a binary classification problem: orthographic schwas are either fully retained or fully deleted when spoken. Previous work has shown that even with rich linguistic representations of words, it is difficult to discover categorical rules that can predict schwa deletion. This led us to approach the problem with machine learning, which we felt would stand a better chance at attaining high performance.

We obtained training data from digitized dictionaries hosted by the University of Chicago \href{https://dsalsrv04.uchicago.edu/dictionaries/}{Digital Dictionaries of South Asia} project. The Hindi data, comprised of the original Devanagari orthography and the phonemic transcription, was parsed out of \citet{mcgregor} and \citet{bahri} and transcribed into an ASCII format. The Punjabi data was similarly processed from \citet{singh}. \Cref{table:entry-example} gives an example entry from the \citeauthor{mcgregor} Hindi dataset.
    
To find all instances of schwa retention and schwa deletion, we force-aligned orthographic and phonemic representations of each dictionary entry using a linear-time algorithm. In cases where force-alignment failed due to idiosyncrasies in the source data (typos, OCR errors, etc.)\ we discarded the entire word. We provide statistics about our datasets in \cref{table:datasets}. We primarily used the dataset from \citeauthor{mcgregor} in training our Hindi models due to its comprehensiveness and high quality.

\begin{table}[t!]
\small
\begin{center}
\begin{tabular}{| r | l |}
    \hline
    \textbf{Devanagari} & {\dn a\1kwAhV} \\
    \textbf{Orthographic} & \texttt{a \textasciitilde{} k a rr aa h a tt a} \\
    \textbf{Phonemic} & \texttt{a \textasciitilde{} k\,\,\,\,\, rr aa h a tt} \\
    \hline
\end{tabular}
\end{center}
\caption{\label{table:entry-example} An example entry from the Hindi training dataset. }
\end{table}

\begin{table}[t!]
\small
\begin{center}
\begin{tabular}{|r|rrr|}
    \hline
    \textbf{Hindi Dict.} & \textbf{Entries} & \textbf{Schwas} & \textbf{Deletion Rate} \\
    \hline
    McGregor & 34,952 & 36,183 & 52.94\% \\
    Bahri & 9,769 & 14,082 & 49.41\% \\
    Google & 847 & 1,098 & 56.28\% \\
    \hline\hline
    \textbf{Punjabi Dict.} & \textbf{Entries} & \textbf{Schwas} & \textbf{Deletion Rate} \\\hline
    Singh & 28,324 & 34,576 & 52.25\% \\
    \hline
\end{tabular}
\end{center}
\caption{\label{table:datasets} Statistics about the datasets used. The deletion rate is the percentage of schwas that are deleted in their phonemic representation. The Google dataset, taken from \citet{johny_brahmic_2018}, was not considered in our final results due to its small size and over-representation of proper nouns.}
\end{table}

Each schwa instance was an input in our training set. The output was a boolean value indicating whether the schwa was retained. Our features in the input column were a one-hot encoding of a variable window of phones to the left ($c_{-n}, \dots, c_{-1}$) and right ($c_{+1}, \dots, c_{+m}$) of the schwa instance ($c_0$) under consideration. The length of the window on either side was treated as a hyperparamater and tuned. We also tested whether including phonological features (for vowels: height, backness, roundedness, and length; for consonants: voice, aspiration, and place of articulation) of the adjacent graphemes affected the accuracy of the model.

We trained three models on each dataset: logistic regression from scikit-learn, MLPClassifier (multilayer perceptron neural network) from scikit-learn, and XGBClassifier (gradient-boosting decision trees) from XGBoost. We varied the size of the window of adjacent phonemes and trained with and without phonological feature data.

\section{Results}

\begin{table}[t!]
\small
\begin{center}
\begin{tabular}{|r|llll|l|}
    \hline
    \textbf{} & \multicolumn{1}{c}{\textbf{Model}} & \multicolumn{1}{c}{\textbf{A}} & \multicolumn{1}{c}{\textbf{P}} & \multicolumn{1}{c|}{\textbf{R}} & \multicolumn{1}{c|}{\textbf{Word A}} \\
    \hline
    Hindi & XGBoost & 98.00\% & 98.04\% & 97.60\% & 97.78\% \\
    & Neural & 97.83\% & 97.86\% & 97.42\% & 97.62\% \\
    & Logistic & 97.19\% & 97.19\% & 96.70\% & 96.86\% \\
    & \citeauthor{wiktionary} & 94.18\% & 92.89\% & 94.29\% & 94.18\% \\
    \hline
    Punjabi & XGBoost & 94.66\% & 92.79\% & 95.90\% & 94.18\% \\
    & Neural & 94.66\% & 93.25\% & 95.47\% & 94.07\% \\
    & Logistic & 93.77\% & 91.73\% & 95.04\% & 93.14\% \\
    \hline
\end{tabular}
\end{center}
\caption{\label{table:results} Results for our models on the \citeauthor{mcgregor} and \citeauthor{singh} datasets: Per-schwa accuracy, precision, and recall, as well as word-level accuracy (all schwas in the word must be correctly classified).} 
\end{table}

\Cref{table:results} tabulates the performances of our various models.

We obtained a maximum of 98.00\% accuracy for all schwa instances in our test set from the McGregor dataset with gradient-boosted decision trees from XGBoost. We used a window of 5 phonemes to the left and right of the schwa instance, phonological features, 200 estimators, and a maximum tree depth of 11. Any model with at least 200 estimators and a depth of at least 5 obtains a comparable accuracy, but this gradually degrades with increasing estimators due to overfitting. Without phonological feature data, the model consistently achieves a slightly lower accuracy of 97.93\%.

Logistic regression with the same features achieved 97.19\% accuracy. An MLP classifier with a single hidden layer of 250 neurons and a learning rate of $10^{-4}$ achieved 97.83\% accuracy.

On the Singh dataset for Punjabi, the same XGBoost model (except without phonological features) achieved 94.66\% accuracy. This shows the extensibility of our system to other Indo-Aryan languages that undergo schwa deletion.

We were unable to obtain evaluation datasets or code from previous work (\citealt{narasimhan_schwa-deletion_2004}, \citealt{tyson_prosodic_2009}) for a direct comparison of our system with previous ones.\footnote{We were able to obtain code from \citet{roy_2017} but were unable to run it on our machines.} However, we were able to port and test the Hindi transliteration code written in Lua utilized by \citet{wiktionary}, an online freely-editable dictionary operated by the Wikimedia Foundation, the parent of Wikipedia. That system obtains 94.94\% word-level accuracy on the \citeauthor{mcgregor} dataset, which we outperform consistently.

\section{Discussion}


Our system achieved higher performance than any other.

The schwa instances which our model did not correctly predict tended to fall into two classes: borrowings from Persian, Arabic, or European languages, or compounds of native or Sanskrit-borrowed morphemes. Of the 150 Hindi words from our test set from \citeauthor{mcgregor} that our best model incorrectly predicted schwa deletion for, we sampled 20 instances and tabulated their source languages. 10 were native Indo-Aryan terms descended through the direct ancestors of Hindi, 4 were learned Sanskrit borrowings, 5 were Perso-Arabic borrowings, and 1 was a Dravidian borrowing. 9 were composed of multiple morphemes. Borrowings are overrepresented relative to the baseline rate for Hindi; in one frequency list, only 8 of the 1,000 top words in Hindi were of Perso-Arabic origin (\citealt{ghatage}).

Notably, some of the Perso-Arabic borrowings that the model failed on actually reflected colloquial pronunciation; e.g.~{\dn amn} \ortho{\ipa{@m@n@}} is \ipa{[@mn]} in \citeauthor{mcgregor} yet our model predicts \ipa{[@m@n]} which is standard in most speech.

We qualitatively analyzed our system to investigate what kind of linguistic representations it seemed to be learning. To do this, we inspected several decision trees generated in our model, and found that our system was learning both prosodic and phonetic patterns.

Some trees very clearly encoded phonotactic information. One tree we examined had a subtree that could be paraphrased like so, where $c_n$ indicates the phone $n$ characters away from the schwa being considered: ``If $c_{+1}$ is beyond the end of the word, and $c_{-2}$ is not beyond the beginning of the word, and $c_{-2}$ is a \oipa{t}, then if $c_{-1}$ is a \oipa{j}, then penalize deleting this schwa;\footnote{{\it Penalize deleting} and not {\it delete}, because this tree is only contributing towards the final decision, along with all the other trees.} otherwise if $c_{-1}$ is not a \oipa{j}, prefer deleting this schwa.'' Put another way, this subtree penalizes deleting a schwa if it comes at the end of a word, the preceding two characters are exactly \oipa{tj}, and the word extends beyond the preceding two characters. This is just the kind of phonetic rule that systems like \citet{narasimhan_schwa-deletion_2004} were using.

The extent to which our system encodes prosodic information was less clear. Our features were phonetic, not prosodic, but some prosodic information can be somewhat captured in terms of phonetics. Take, for instance, this subtree that we found in our model, paraphrasing as before: 
``If $c_{-3}$ is beyond the beginning of the word, and $c_{-2}$ is \oipa{a:}, then if $c_{+2}$ is \oipa{@}, prefer deletion; otherwise, if $c_{+2}$ is not \oipa{@}, penalize deletion.'' 
Consider this rule as it would apply to the first schwa in the Hindi word {\dn aAmdnF}
\ortho{\ipa{a:m@d@ni:}}

\vspace{0.4em}
\begin{center}
\begin{tabular}{cccccccccc}
    -3 & -2 & -1 & \textbf{0} & 1 & 2 & 3 & 4 & 5 \\
    & \textipa{a:} & \textipa{m} & \textbf{\ipa{@}} & \textipa{d} & \ipa{@} & \textipa{n} & \ipa{i:}
\end{tabular}
\end{center}
\vspace{0.4em}

\noindent The rule decides that deleting the first schwa should be penalized, and it decided this by using criteria that entail that the preceding rhyme is heavy and the following rhyme is light.\footnote{Actually, this is not exactly true, since if the following syllable had any consonants in the rhyme, it would become heavy, even if there were a schwa present. But this is an error that could be corrected by other decision trees.} Obviously, though, this same rule would not work for other heavy and light syllables: if any of the vowels had been different, or at different offsets, a non-deletion rather than a deletion would have been preferred, which is not what it ought to do if it is emulating the prosodic rule. 

It is expected that our model is only able to capture ungeneralized, low-level patterns like this, since it lacks the symbolic vocabulary to capture elegant linguistic generalizations, and it is perhaps surprising that our system is able to achieve the performance it does even with this limitation. In future work, it would be interesting to give our system more directly prosodic representations, like the moraic weights of the surrounding syllables and syllabic stress. 
    
Another limitation of our system is that it assumes all schwas are phonologically alike, which may not be the case. While most schwas are at all times either pronounced or deleted, there are less determinate cases where a schwa might or might not be deleted according to sociolinguistic and other factors.  \Citet[p.~xi]{mcgregor} calls these ``weakened schwas'', describing them as ``weakened by Hindi speakers in many phonetic contexts, and dropped in others'' and orthographically indicating them with a breve. {\dn s(y} is transcribed \textit{satyă}. Our best model correctly classified 80.4\% of the weakened schwas present in our test set taken from \mbox{McGregor}. Improving our performance for this class of schwas may require us to treat them differently from other schwas. Further research is needed on the nature of weakened schwas.

\section{Conclusion}

We have presented the first statistical schwa deletion classifier for Hindi achieves state-of-the-art performance. Our system requires no hard-coded phonological rules, instead relying solely on pairs of orthographic and phonetic forms for Hindi words at training time.

Furthermore, this research presents the first schwa-deletion model for Punjabi, and has contributed several freely-accessible scripts for scraping Hindi and Punjabi pronunciation data from online sources.

\normalfont 
\bibliographystyle{acl_natbib}
\bibliography{acl2020}

\end{document}